\documentclass{article}

\usepackage{arxiv}

\usepackage[utf8]{inputenc} % allow utf-8 input
\usepackage[T1]{fontenc}    % use 8-bit T1 fonts
\usepackage{hyperref}       % hyperlinks
\usepackage{url}            % simple URL typesetting
\usepackage{booktabs}       % professional-quality tables
\usepackage{amsfonts}       % blackboard math symbols
\usepackage{nicefrac}       % compact symbols for 1/2, etc.
\usepackage{microtype}      % microtypography
\usepackage{lipsum}		% Can be removed after putting your text content
\usepackage{graphicx}
\usepackage{natbib}
\usepackage{doi}
\usepackage[skins, breakable]{tcolorbox}
\usepackage{listings}
\usepackage{xcolor}
\usepackage{amsmath}
\usepackage{algorithm}
\usepackage{algorithmic}
\usepackage{amssymb}
\title{LLMize: A Framework for Large Language Model-Based Numerical Optimization\thanks{Source code available at \url{https://github.com/rizkiokt/llmize}.}}

\author{ 
    \href{https://orcid.org/0000-0003-2814-1991}{\includegraphics[scale=0.06]{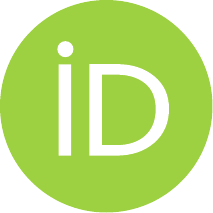}\hspace{1mm}M. Rizki Oktavian}$^{1,2}$ \\ 
    \\
    \textsuperscript{}{rizki@bwailabs.com} \\
    \textsuperscript{1}Blue Wave AI Labs, West Lafayette, IN, United States\\
    \textsuperscript{2}School of Nuclear Engineering, Purdue University, West Lafayette, IN, United States
}
\date{}

\renewcommand{\headeright}{}
\renewcommand{\undertitle}{}
\renewcommand{\shorttitle}{LLMize: A Framework for Large Language Model-Based Numerical Optimization}

\hypersetup{
pdftitle={Optimization through Iterative LLM Prompting for Nuclear Engineering Design Problems},
pdfsubject={cs.LG, physics.comp-ph},
pdfauthor={M. Rizki Oktavian},
pdfkeywords={Optimization, In-Context Learning,  Large Language Models},
}

\begin{document}
\maketitle

\begin{abstract}
Large language models (LLMs) have recently shown strong reasoning capabilities beyond traditional language tasks, motivating their use for numerical optimization. This paper presents \textit{LLMize}, an open-source Python framework that enables LLM-driven optimization through iterative prompting and in-context learning. LLMize formulates optimization as a black-box process in which candidate solutions are generated in natural language, evaluated by an external objective function, and refined over successive iterations using solution--score feedback. The framework supports multiple optimization strategies, including Optimization by Prompting (OPRO) and hybrid LLM-based methods inspired by evolutionary algorithms and simulated annealing. A key advantage of LLMize is the ability to inject constraints, rules, and domain knowledge directly through natural language descriptions, allowing practitioners to define complex optimization problems without requiring expertise in mathematical programming or metaheuristic design. LLMize is evaluated on convex optimization, linear programming, the Traveling Salesman Problem, neural network hyperparameter tuning, and nuclear fuel lattice optimization. Results show that while LLM-based optimization is not competitive with classical solvers for simple problems, it provides a practical and accessible approach for complex, domain-specific tasks where constraints and heuristics are difficult to formalize.
 
\end{abstract}

% keywords can be removed

\section{Introduction}

Numerical optimization is a foundational component of scientific computing, machine learning, and engineering design. Many real-world optimization problems are inherently black-box in nature, where objective functions are evaluated through simulations, experiments, or complex pipelines rather than closed-form expressions. In such settings, gradients are unavailable, constraints are implicit, and evaluations are often expensive. Examples include hyperparameter tuning, combinatorial optimization, and engineering design problems driven by physics-based simulators. While classical optimization methods such as gradient-based solvers, evolutionary algorithms, and Bayesian optimization are effective when mathematical structure is available, they become difficult to apply when objectives and constraints are hard to formalize.

Recent advances in large language models (LLMs) suggest a new direction for black-box optimization. Although LLMs were originally developed for natural language understanding and generation, they exhibit strong in-context learning capabilities. In-context learning allows a model to adapt its behavior based on examples and feedback provided directly in the prompt, without updating model parameters \citep{brown2020language}. This capability enables LLMs to recognize patterns, reason over historical information, and refine outputs across iterative interactions. As a result, LLMs have been explored as general-purpose reasoning engines beyond traditional language tasks \citep{wei2022chain}.

In optimization contexts, in-context learning enables an LLM to observe previously evaluated solutions and their objective values, and to propose improved candidates conditioned on this history. This approach does not rely on gradients, surrogate models, or hand-crafted search operators. Instead, problem structure, constraints, and heuristics can be communicated directly through natural language descriptions. This makes LLM-based optimization particularly attractive for domains where expert knowledge exists but is difficult to encode mathematically.

A representative example of this paradigm is Optimization by PROmpting (OPRO), introduced by \citet{yang2024opro}. OPRO formulates optimization as an iterative prompting process, where an LLM generates candidate solutions based on a problem description and a set of historical solution--score pairs. After each evaluation, the prompt is updated with feedback, allowing the LLM to refine future proposals through in-context learning. OPRO demonstrated that LLMs can achieve competitive performance in low-budget optimization scenarios, including prompt optimization and small-scale numerical and combinatorial problems.

Building on this idea, recent studies have explored LLMs for optimization-related tasks such as algorithmic reasoning, combinatorial search, and hyperparameter tuning. These works suggest that LLMs can combine prior knowledge learned during pretraining with feedback from external evaluators to guide search effectively. However, most existing approaches remain task-specific or focus on isolated demonstrations. A general, extensible framework that integrates LLM-driven proposal generation with established optimization principles remains limited.

This paper introduces \textbf{LLMize}, an open-source Python framework for numerical optimization using large language models. LLMize generalizes the OPRO paradigm into a practical and reusable optimization framework. It implements three complementary optimization strategies: OPRO-style iterative prompting, a Hybrid LLM--Evolutionary Algorithm (HLMEA), and a Hybrid LLM--Simulated Annealing method (HLMSA). In all cases, the LLM functions as a proposal generator, while the optimization loop explicitly controls evaluation, selection, and termination.

A central strength of LLMize is the ability to inject constraints, rules, and domain knowledge directly through natural language. Instead of encoding heuristics as mathematical constraints or custom operators, users can describe objectives, safety limits, and design preferences in plain English. This design enables practitioners to apply optimization techniques without requiring expertise in mathematical programming or metaheuristic algorithm design. The optimization logic emerges from in-context learning over historical solutions rather than fixed update rules.

LLMize is model-agnostic and problem-agnostic. Optimization problems are defined using a textual description and a user-provided black-box objective function. The framework manages prompt construction, candidate generation, batched evaluation, adaptive parameter control, and result tracking. This design supports a wide range of applications, including convex optimization \citep{boyd2004convex}, combinatorial optimization \citep{garey1979computers}, machine learning hyperparameter tuning \citep{bergstra2012random}, and physics-based engineering design.

The practical value of this approach is demonstrated through multiple case studies, including engineering optimization. In prior work, the author applied OPRO to Boiling Water Reactor fuel lattice design and showed that LLM-based optimization can match or outperform a genetic algorithm while allowing reactor physics constraints and heuristics to be expressed directly in natural language \citep{oktavian2024nuclearopro}. This result highlights the potential of LLM-driven optimization for complex, safety-critical engineering problems where constraints are highly structured but difficult to formalize mathematically.

The main contributions of this work are summarized as follows:
\begin{itemize}
    \item Introduction of \textbf{LLMize}, an open-source framework for LLM-driven black-box numerical optimization.
    \item Integration of in-context learning with classical optimization ideas through OPRO, HLMEA, and HLMSA.
    \item A unified interface supporting constraint injection, batched evaluation, adaptive control, and callback-based termination.
    \item Empirical evaluation across numerical, combinatorial, machine learning, and engineering optimization problems.
\end{itemize}

LLMize is not intended to replace classical optimization algorithms in settings where gradients, convexity, or exact solvers are available. Instead, it targets complex, domain-specific problems where objectives and constraints are difficult to formalize and expert knowledge is most naturally expressed in language. By framing optimization as an iterative process guided by in-context learning, LLMize demonstrates how large language models can serve as flexible and accessible black-box optimizers.

\section{Methodology} 
\subsection{Problem Formulation}

This work considers numerical optimization problems that can be evaluated through a black-box objective function. Let $\mathcal{X}$ denote the search space of candidate solutions, where each solution $x \in \mathcal{X}$ is represented in a form that can be expressed as text or structured data and parsed programmatically. The optimization objective is defined as
\begin{equation}
    \min_{x \in \mathcal{X}} \; f(x)
    \quad \text{or} \quad
    \max_{x \in \mathcal{X}} \; f(x),
\end{equation}
where $f(x)$ is evaluated by an external process, such as a numerical routine, simulation, or trained model. No assumptions are made regarding differentiability, convexity, or continuity of $f(x)$, which places the problem in the category of black-box optimization \citep{audet2017derivativefree}.

Constraints may be present in the form of equality or inequality conditions. These constraints are enforced implicitly through the objective evaluation by penalizing infeasible solutions, or explicitly described as part of the problem specification. Let $\mathcal{X}_{\text{feas}} \subseteq \mathcal{X}$ denote the feasible region. In practice, feasibility is determined by the objective evaluation rather than analytical constraint functions, which is common in simulator-based optimization \citep{forrester2009recent}.

In LLMize, a large language model is used as a proposal mechanism rather than a direct evaluator. At iteration $t$, the language model generates candidate solutions conditioned on a prompt that contains the problem description and a history of previously evaluated solutions and their objective values. This process can be expressed as sampling from a conditional distribution
\begin{equation}
    x_t \sim p_{\theta}(x \mid \mathcal{H}_t),
\end{equation}
where $\theta$ denotes the fixed parameters of the language model and
\begin{equation}
    \mathcal{H}_t = \{(x_1, f(x_1)), (x_2, f(x_2)), \dots, (x_{t-1}, f(x_{t-1}))\}
\end{equation}
represents the optimization history available in the prompt. The language model parameters are not updated during optimization. Adaptation occurs entirely through conditioning on $\mathcal{H}_t$, which corresponds to in-context learning \citep{brown2020language}.

After candidate generation, each proposed solution is evaluated using the objective function, and the history is updated as
\begin{equation}
    \mathcal{H}_{t+1} = \mathcal{H}_t \cup \{(x_t, f(x_t))\}.
\end{equation}
The optimization process proceeds iteratively until a termination condition is met, such as a fixed evaluation budget or lack of improvement over multiple iterations. This formulation is closely related to prompt-based optimization methods, including Optimization by PROmpting (OPRO) \citep{yang2024opro}.

A central advantage of this formulation is the ability to inject domain knowledge directly into the optimization process using natural language. Problem constraints, heuristics, preferred solution structures, and expert intuition can be included in the textual problem description without requiring mathematical reformulation. This allows domain experts to guide the optimization process using plain English explanations, which are incorporated by the language model through in-context learning.

\subsection{OPRO-Based Optimization in LLMize}

The first optimization strategy implemented in LLMize follows the Optimization by PROmpting (OPRO) paradigm \citep{yang2024opro}. In this approach, the large language model serves as an iterative proposal generator that refines candidate solutions based on objective feedback provided through the prompt.

At iteration $t$, a prompt $\mathcal{P}_t$ is constructed from three components: (i) a textual description of the optimization problem, (ii) optional domain knowledge or heuristics expressed in natural language, and (iii) the optimization history $\mathcal{H}_t$. Formally, the prompt can be written as
\begin{equation}
    \mathcal{P}_t = \texttt{Prompt}(\text{Problem}, \text{Domain Knowledge}, \mathcal{H}_t).
\end{equation}

Given this prompt, the language model generates one or more candidate solutions by sampling from its conditional distribution,
\begin{equation}
    x_t \sim p_{\theta}(x \mid \mathcal{P}_t),
\end{equation}
where $\theta$ denotes the fixed parameters of the language model. The generated output is parsed into a structured solution representation compatible with the objective evaluation function.

Each proposed candidate is then evaluated using the black-box objective function $f(x_t)$. The resulting score is appended to the optimization history, producing
\begin{equation}
    \mathcal{H}_{t+1} = \mathcal{H}_t \cup \{(x_t, f(x_t))\}.
\end{equation}
This updated history is included in the prompt for the next iteration, enabling the language model to adapt its future proposals through in-context learning \citep{brown2020language}.

Unlike gradient-based or model-based optimization methods, OPRO does not rely on explicit search operators or surrogate models. Instead, improvement emerges from the language model’s ability to recognize patterns in previously evaluated solutions and to propose new candidates that are likely to improve the objective. This behavior resembles heuristic search guided by learned priors rather than explicit optimization rules.

In LLMize, the OPRO process is extended to support both minimization and maximization objectives, batched candidate generation, and parallel evaluation. When multiple candidates are generated at iteration $t$, the history update generalizes to
\begin{equation}
    \mathcal{H}_{t+1} = \mathcal{H}_t \cup \{(x_t^{(i)}, f(x_t^{(i)}))\}_{i=1}^{B},
\end{equation}
where $B$ denotes the batch size. This allows more efficient exploration of the search space under a fixed evaluation budget.

The OPRO-based optimization process can be summarized as an iterative loop that alternates between candidate generation using a language model and objective-based evaluation using an external oracle. At each iteration, the prompt is updated to include newly evaluated solutions, allowing the language model to adapt its proposals through in-context learning. This process continues until a termination condition is met, such as a fixed evaluation budget or lack of improvement. Algorithm~\ref{alg:opro} provides a step-by-step description of the OPRO-based optimization procedure as implemented in LLMize.

\begin{algorithm}[H]
\caption{OPRO-Based Optimization in LLMize}
\label{alg:opro}
\begin{algorithmic}[1]
\REQUIRE Problem description $\mathcal{D}$, objective function $f(x)$, language model with parameters $\theta$, initial history $\mathcal{H}_1$, maximum iterations $T$, batch size $B$
\ENSURE Best solution $x^{*}$ and corresponding objective value $f(x^{*})$

\STATE Initialize best solution $x^{*} \leftarrow \varnothing$
\STATE Initialize best score $f(x^{*}) \leftarrow -\infty$ (or $+\infty$ for minimization)

\FOR{$t = 1$ to $T$}
    \STATE Construct prompt $\mathcal{P}_t \leftarrow \texttt{Prompt}(\mathcal{D}, \mathcal{H}_t)$
    \STATE Generate candidate solutions $\{x_t^{(i)}\}_{i=1}^{B} \sim p_{\theta}(x \mid \mathcal{P}_t)$
    
    \FOR{each candidate $x_t^{(i)}$}
        \STATE Evaluate objective value $y_t^{(i)} \leftarrow f(x_t^{(i)})$
        \STATE Append $(x_t^{(i)}, y_t^{(i)})$ to history $\mathcal{H}_t$
        
        \IF{$y_t^{(i)}$ improves upon $f(x^{*})$}
            \STATE Update best solution $x^{*} \leftarrow x_t^{(i)}$
            \STATE Update best score $f(x^{*}) \leftarrow y_t^{(i)}$
        \ENDIF
    \ENDFOR
    
    \STATE Update history $\mathcal{H}_{t+1} \leftarrow \mathcal{H}_t$
\ENDFOR

\RETURN $x^{*}, f(x^{*})$
\end{algorithmic}
\end{algorithm}

\subsection{HLMEA: Hybrid LLM--Evolutionary Algorithm}

The second optimization strategy implemented in LLMize is a Hybrid LLM--Evolutionary Algorithm (HLMEA). HLMEA is a population-based, LLM-driven optimization method inspired by evolutionary algorithms and recent Language Model Evolutionary Algorithms (LMEA) \citep{liu2023lmea}. Unlike classical evolutionary algorithms, HLMEA does not implement mutation, crossover, or selection operators explicitly. Instead, these operations are delegated to a large language model through structured natural language prompts.

At iteration $t$, HLMEA generates a batch of candidate solutions,
\begin{equation}
    \mathcal{P}_t = \{x_t^{(1)}, x_t^{(2)}, \dots, x_t^{(B)}\},
\end{equation}
where $B$ is the batch size. Each candidate is evaluated using the black-box objective function $f(x)$, producing a corresponding set of objective values. No assumptions are made about the structure of the objective function, making HLMEA suitable for black-box and simulator-based optimization problems.

To guide the generation of new candidates, HLMEA maintains a truncated history of solution--score pairs,
\begin{equation}
    \mathcal{H}_t = \{(x_i, f(x_i))\}_{i=1}^{K},
\end{equation}
where $K$ is a fixed history size. This history is constructed by retaining the highest-performing solutions observed so far. Elitism is therefore implemented implicitly by preserving strong solutions in the prompt context rather than explicitly copying individuals across generations.

At each iteration, a prompt is constructed that includes the problem description, the truncated optimization history $\mathcal{H}_t$, and optional domain knowledge expressed in natural language. The language model is instructed to reason as an evolutionary algorithm by selecting parents, applying crossover and mutation strategies, and enforcing solution uniqueness. New candidates are generated according to
\begin{equation}
    x_{t+1}^{(i)} \sim p_{\theta}(x \mid \mathcal{H}_t),
\end{equation}
where $\theta$ denotes the fixed parameters of the language model. Adaptation occurs through in-context learning, as the model conditions on past solutions and objective feedback without updating its parameters \citep{brown2020language}.

In addition to proposing candidate solutions, the language model is prompted to select evolutionary hyperparameters, including elitism rate, mutation rate, and crossover rate. These hyperparameters are not enforced algorithmically but influence the reasoning process used by the model during candidate generation. This design allows HLMEA to function as a hyper-heuristic optimizer, where evolutionary behavior emerges from language-guided reasoning rather than fixed operators.

HLMEA proceeds iteratively until a termination condition is met, such as a fixed number of steps or callback-based early stopping. By expressing evolutionary logic and domain heuristics directly in natural language, HLMEA enables flexible population-based optimization while avoiding problem-specific operator design. Algorithm \ref{alg:hlmea} explain this process in details.

\begin{algorithm}[H]
\caption{HLMEA: LLM-Driven Evolutionary Optimization}
\label{alg:hlmea}
\begin{algorithmic}[1]
\REQUIRE Problem description $\mathcal{D}$, objective function $f(x)$, language model with parameters $\theta$, initial examples $\mathcal{H}_1$, batch size $B$, maximum iterations $T$
\ENSURE Best solution $x^{*}$ and objective value $f(x^{*})$

\STATE Initialize optimization history $\mathcal{H}_1$ using initial solutions and scores
\STATE Initialize best solution $x^{*}$ from $\mathcal{H}_1$

\FOR{$t = 1$ to $T$}
    \STATE Construct prompt $\mathcal{P}_t \leftarrow \texttt{Prompt}(\mathcal{D}, \mathcal{H}_t)$
    \STATE Generate $B$ candidate solutions $\{x_t^{(i)}\}_{i=1}^{B} \sim p_{\theta}(x \mid \mathcal{P}_t)$
    
    \FOR{each candidate $x_t^{(i)}$}
        \STATE Evaluate $f(x_t^{(i)})$
        \IF{$f(x_t^{(i)})$ improves upon $f(x^{*})$}
            \STATE Update best solution $x^{*} \leftarrow x_t^{(i)}$
        \ENDIF
    \ENDFOR
    
    \STATE Update history $\mathcal{H}_{t+1}$ by retaining top-performing solution--score pairs
\ENDFOR

\RETURN $x^{*}, f(x^{*})$
\end{algorithmic}
\end{algorithm}

\subsection{HLMSA: Hybrid LLM--Simulated Annealing}

The third optimization strategy implemented in LLMize is a Hybrid LLM--Simulated Annealing method (HLMSA). This approach combines the structure of classical simulated annealing with LLM-driven candidate generation. Simulated annealing is a well-established metaheuristic for black-box optimization and is effective at escaping local optima through controlled stochastic acceptance of suboptimal solutions \citep{kirkpatrick1983optimization}.

HLMSA operates on a small batch of parallel solution trajectories that share a global temperature schedule. At iteration $t$, each trajectory maintains a current solution $x_t^{(i)}$ and corresponding objective value $f(x_t^{(i)})$. A temperature parameter $T_t > 0$ controls the level of stochastic exploration across all trajectories.

At each iteration, a prompt is constructed that includes the problem description, examples of previously evaluated solutions and their objective values, and optional domain knowledge expressed in natural language. The language model then generates a batch of candidate solutions conditioned on this information,
\begin{equation}
    x'^{(i)} \sim p_{\theta}(x \mid \mathcal{H}_t),
\end{equation}
where $\mathcal{H}_t$ denotes the optimization history included in the prompt. The language model parameters remain fixed throughout optimization. Adaptation occurs through in-context learning, as the model conditions on observed solutions and outcomes \citep{brown2020language}. This formulation is consistent with prior work that frames large language models as feedback-driven proposal generators for search and optimization \citep{yao2023react,shinn2023reflexion}.

Each proposed candidate is evaluated using the objective function. For each trajectory, improving candidates are always accepted. Worse candidates may still be accepted with a probability that decreases with both the objective difference and the current temperature. This acceptance mechanism enables early exploration and later exploitation without requiring gradient information.

After each iteration, the temperature is updated according to
\begin{equation}
    T_{t+1} = \alpha_t T_t,
\end{equation}
where the cooling rate $\alpha_t \in (0,1)$ is selected dynamically. Unlike classical simulated annealing, the cooling rate in HLMSA is proposed by the language model itself and parsed from the generated output. This allows the annealing schedule to adapt based on optimization progress rather than following a fixed predefined rule.

HLMSA differs from classical simulated annealing primarily in how neighboring solutions are generated and how the cooling schedule is controlled. Instead of relying on problem-specific perturbation operators, HLMSA uses a language model to propose informed local modifications guided by prior solutions, objective feedback, and injected domain knowledge. Constraints, heuristics, and preferred solution structures can be provided directly in plain English, without modifying the optimization algorithm.

\subsection{Implementation Details}

LLMize is implemented as a modular Python framework designed to support flexible integration of large language models into numerical optimization workflows. All optimizers inherit from a common \texttt{Optimizer} base class, which defines a shared interface for problem specification, solution generation, objective evaluation, and result tracking. This design allows different optimization strategies to share infrastructure while maintaining algorithm-specific logic. Table~\ref{tab:implementation-details} summarizes the key framework components and implementation features provided by LLMize.

\begin{table}[H]
\centering
\caption{Implementation Features of the LLMize Framework}
\label{tab:implementation-details}
\begin{tabular}{p{3.5cm} p{10.5cm}}
\hline
\textbf{Component} & \textbf{Description} \\
\hline
Optimizer Base Class &
All optimization strategies inherit from a common \texttt{Optimizer} interface, which defines shared logic for prompt construction, solution generation, objective evaluation, and result tracking. \\

Problem Specification &
Optimization problems are defined using a natural language description and a user-provided black-box objective function. \\

Prompt Management &
Prompts are dynamically constructed using the problem description and a truncated history of solution--score pairs. \\

Solution Parsing &
Generated outputs are parsed into structured solution representations using lightweight parsing utilities. \\

Batch Generation &
Each iteration can generate multiple candidate solutions in parallel. Batch size is configurable and shared across optimization strategies. \\

Parallel Evaluation &
Objective function evaluations can be parallelized across multiple workers, enabling efficient optimization for computationally expensive simulations. \\

Callback System &
A modular callback mechanism supports early stopping, target-based termination, and adaptive control of sampling parameters without modifying optimizer code. \\

Adaptive Control &
Sampling parameters such as model temperature or cooling schedules can be adjusted dynamically based on optimization progress through callbacks or LLM-selected hyperparameters. \\

Model Abstraction &
LLMize is model-agnostic and supports multiple language model providers through a unified initialization interface. \\

Result Tracking &
Optimization results are stored in a structured \texttt{OptimizationResult} object, including best solutions, score histories, and per-iteration statistics. \\
\hline
\end{tabular}
\end{table}

\section{Applications and Case Studies}

The LLMize framework is evaluated across a diverse set of optimization problems to demonstrate its generality and practical applicability. All case studies are formulated as black-box optimization tasks defined by a natural language problem description and a user-provided objective evaluation function. Candidate solutions are generated by a large language model and evaluated iteratively, with previously evaluated solution--score pairs included in the prompt to enable adaptation through in-context learning. Unless otherwise stated, experiments use the OPRO optimization strategy with the instruction-tuned language model \texttt{gemma-3-27b-it}. Early stopping and target-based termination are enabled where applicable to limit unnecessary evaluations. All experiments are executed on a standard workstation, and reported runtimes primarily reflect language model inference overhead.

\subsection{Convex Optimization}

\textbf{Problem.}  
The objective is to minimize the function
\begin{equation}
    f(x_1, x_2) = (x_1 - 3)^2 + (x_2 + 2)^2 + \sin(x_1 + x_2) + 4,
\end{equation}
subject to box constraints
\begin{equation}
    0 \leq x_1 \leq 5, \quad 0 \leq x_2 \leq 5.
\end{equation}
The problem is treated as a black-box optimization task. Constraints are enforced through a large penalty added to the objective for infeasible solutions. Initial samples are generated on a coarse grid over the feasible domain and provided to the optimizer as solution–score pairs.

\begin{figure}[h]
    \centering
    \includegraphics[width=0.6\linewidth]{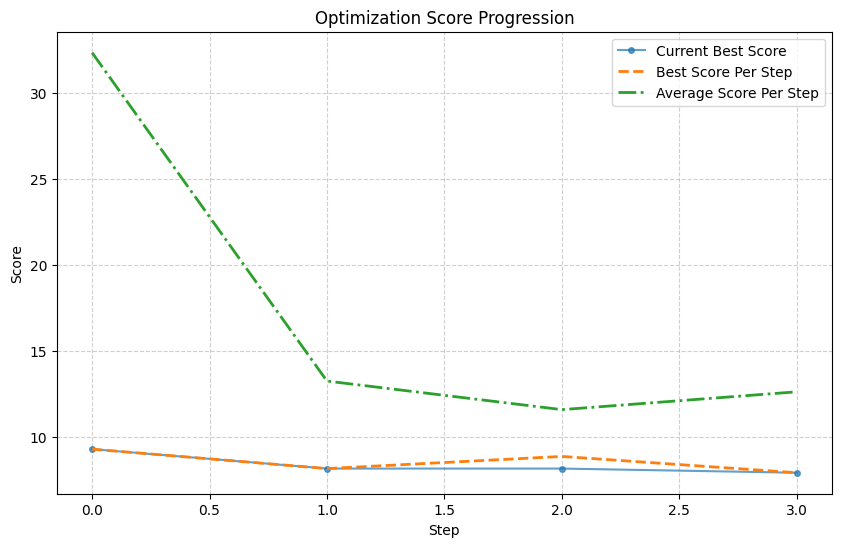}
    \caption{Optimization score progression for the convex optimization problem. The figure shows the current best score, best score per iteration, and average batch score across optimization steps. Lower scores indicate better solutions. A rapid decrease in objective value is observed during early iterations, followed by stabilization once the target threshold is reached.}
    \label{fig:convex-opt}
\end{figure}

\textbf{Results.}  
The optimizer converges rapidly, achieving the predefined target objective value within a few optimization steps. As shown in Figure~\ref{fig:convex-opt}, the best objective value improves sharply after the first iteration and stabilizes near 7.9 by the second to third step. The average batch score decreases significantly during early iterations and remains close to the best score thereafter, indicating reduced variance among proposed solutions. Once the target threshold is satisfied, the optimization terminates early through the optimal-score stopping criterion. The total execution time for the optimization run is approximately 26 seconds, with most of the time spent on language model API calls.

\subsection{Linear Programming}

\textbf{Problem.}  
This case study considers a linear programming maximization problem with three decision variables. The objective is to maximize
\begin{equation}
    Z = 3x_1 + 4x_2 + 6x_3,
\end{equation}
subject to the linear constraints
\begin{align}
    2x_1 + 3x_2 + x_3 &\leq 15, \\
    x_1 + 2x_2 + 3x_3 &\leq 20, \\
    4x_1 + x_2 + 2x_3 &\leq 16, \\
    x_1, x_2, x_3 &\geq 0.
\end{align}
Although this problem admits an exact solution using classical linear programming solvers, it is treated here as a black-box optimization task to evaluate the ability of LLMize to reason over linear constraints using natural language guidance alone.

Constraint handling is implemented using a penalty-based formulation. Candidate solutions that violate any constraint receive a large negative penalty added to the objective value. Initial samples are generated randomly within a broad range and evaluated to provide the optimizer with an initial set of solution--score pairs.

\begin{figure}[h]
    \centering
    \includegraphics[width=0.6\linewidth]{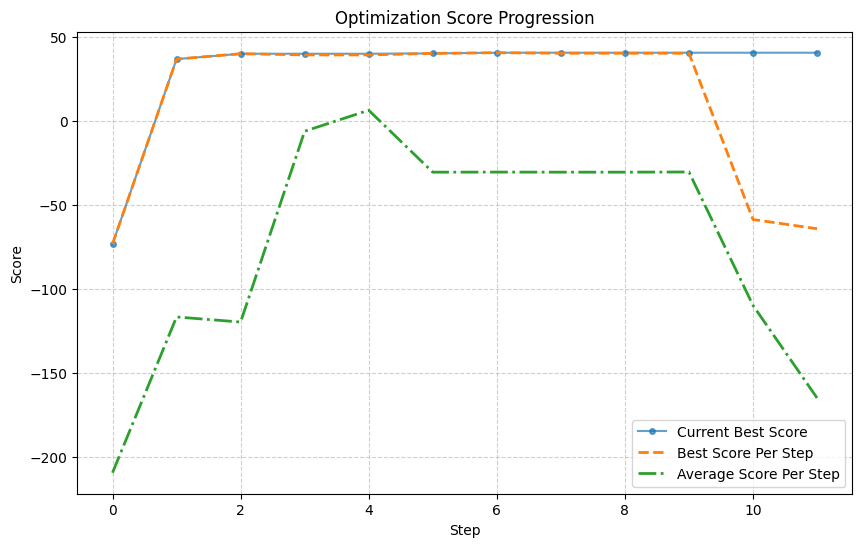}
    \caption{Score progression for the linear programming optimization. The best objective value increases rapidly during early iterations and stabilizes once feasible high-quality solutions are found.}
    \label{fig:lp-opt}
\end{figure}

\textbf{Results.}  
The optimizer exhibits rapid improvement during the early iterations, as shown in Figure~\ref{fig:lp-opt}. Starting from randomly generated initial samples with heavily penalized objective values, the best objective score increases sharply within the first two iterations and stabilizes near 41. This indicates that the language model quickly identifies feasible regions of the solution space and proposes high-quality candidates that satisfy all linear constraints. While the average batch score remains lower due to occasional infeasible proposals receiving penalty values, the current best score remains stable once convergence is reached. The optimization terminates through callback-based stopping criteria after the target objective threshold is achieved, demonstrating effective constraint reasoning and convergence using only black-box evaluations and in-context learning.

\subsection{Traveling Salesman Problem}

\textbf{Problem.}  
This case study considers the Traveling Salesman Problem (TSP) with $N=10$ cities randomly placed in a $100 \times 100$ coordinate space. Each city $i$ has coordinates $(x_i, y_i)$, and the pairwise travel cost is the Euclidean distance
\begin{equation}
    d(i,j) = \sqrt{(x_i - x_j)^2 + (y_i - y_j)^2}.
\end{equation}
A candidate solution is a permutation (route) $\pi$ of $\{0,1,\dots,N-1\}$. The objective is to minimize the closed-tour length
\begin{equation}
    L(\pi) = \sum_{k=0}^{N-2} d(\pi_k, \pi_{k+1}) + d(\pi_{N-1}, \pi_0).
\end{equation}
Invalid routes are assigned a large objective value, enforcing feasibility at evaluation time. Initial routes are generated as random permutations and evaluated to seed the optimization history.

\begin{figure}
    \centering
    \includegraphics[width=0.6\linewidth]{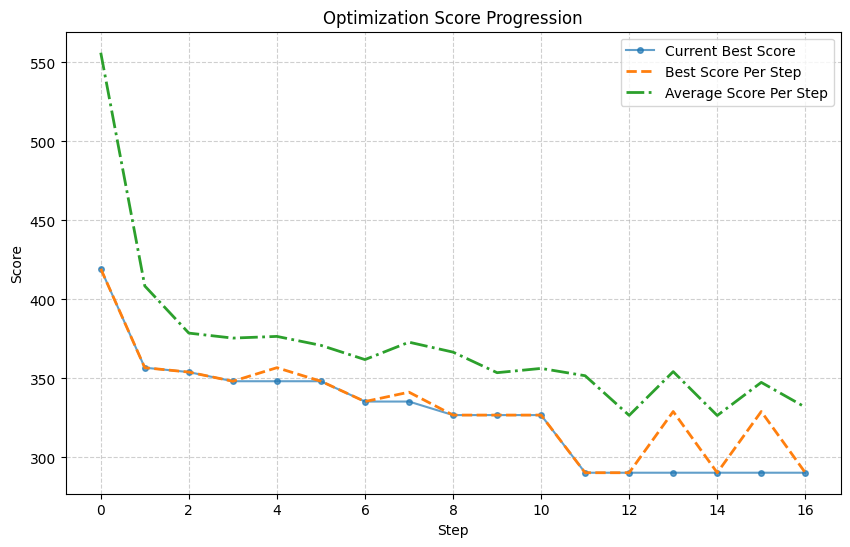}
    \caption{Tour length progression for the TSP case study using OPRO with \texttt{gemini-2.5-flash-lite}. Lower is better.}
    \label{fig:tsp-scores}
\end{figure}

\begin{figure}
    \centering
    \includegraphics[width=0.6\linewidth]{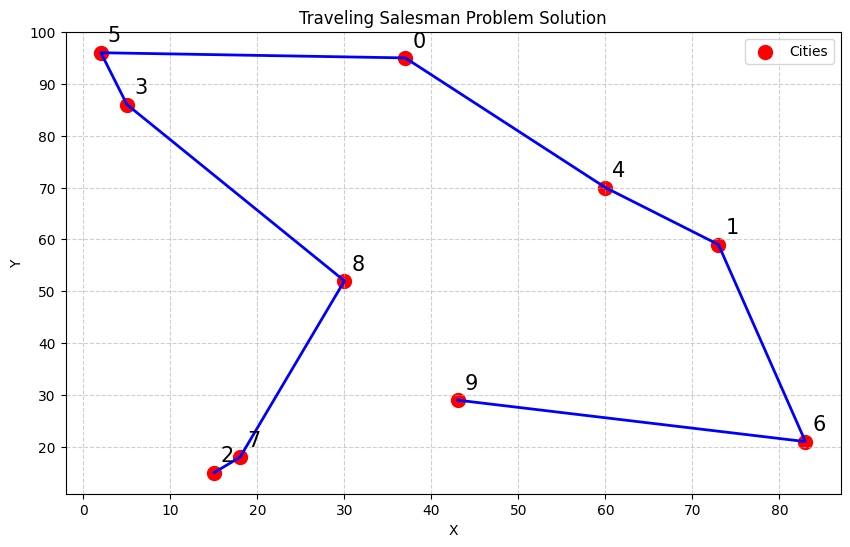}
    \caption{Visualization of the best TSP tour found. Red points indicate city locations, and the blue polyline shows the optimized route connecting all cities in the visiting order returned by the optimizer.}
    \label{fig:tsp-solution}
\end{figure}

\textbf{Results.}  
This experiment uses the OPRO optimizer in LLMize with the language model \texttt{gemini-2.5-flash-lite}, which provides more reliable improvements for this combinatorial task than \texttt{gemma-3-27b-it} under the same evaluation budget. As shown in Figure~\ref{fig:tsp-scores}, the best tour length decreases sharply in the first few steps and continues improving until it stabilizes near 290. The average batch score also decreases over time but remains higher than the best score due to variability across sampled routes. The best tour found in the run is visualized in Figure~\ref{fig:tsp-solution}, demonstrating that LLMize can refine structured permutations through in-context learning and black-box feedback. 

\subsection{Neural Network Hyperparameter Tuning}

\textbf{Problem.}  
This case study evaluates LLMize on neural network hyperparameter tuning using the MNIST handwritten digit classification task. The model architecture consists of a single hidden fully connected layer with ReLU activation, followed by a dropout layer and a softmax output. The optimization variables are the number of hidden units $u$, the dropout rate $p$, and the learning rate $\eta$. For a given hyperparameter configuration $x = (u, p, \eta)$, the objective function trains the model for a fixed number of epochs and returns the validation accuracy,
\begin{equation}
    f(x) = \text{Accuracy}(u, p, \eta).
\end{equation}
The objective is to maximize validation accuracy. Initial hyperparameter configurations are sampled uniformly at random within predefined ranges and evaluated to generate solution--score pairs that seed the optimization history.

\begin{figure}
    \centering
    \includegraphics[width=0.6\linewidth]{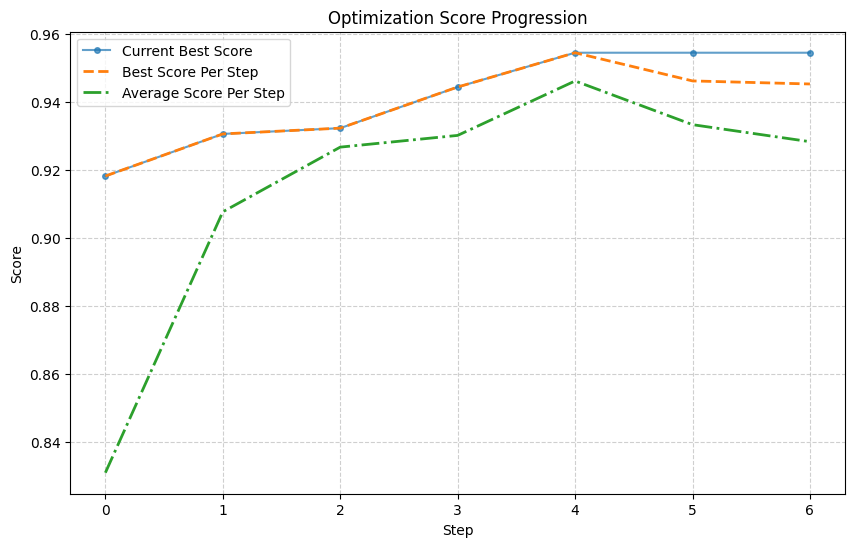}
    \caption{Validation accuracy progression during neural network hyperparameter optimization using OPRO.}
    \label{fig:mnist-hpo}
\end{figure}

\textbf{Results.}  
Using the OPRO optimizer with the instruction-tuned language model \texttt{gemma-3-27b-it}, the best validation accuracy improves steadily over the optimization process as shown in Figure \ref{fig:mnist-hpo}. Starting from an initial best accuracy of 0.918, the optimizer reaches 0.931 after the first iteration and continues improving to 0.945 by step 3. The highest validation accuracy of 0.955 is achieved at step 4, after which no further improvement is observed. Subsequent iterations trigger temperature adaptation and early stopping criteria. The average batch accuracy follows a similar upward trend but remains lower due to variability across sampled hyperparameter configurations. These results indicate that LLMize can efficiently identify high-performing hyperparameter settings within a small number of training evaluations.

\subsection{Nuclear Fuel Lattice Optimization}

\textbf{Problem.}  
This case study presents a nuclear fuel lattice optimization problem as an application of the LLMize framework, building on previous work by the authors \citep{oktavian2024nuclearopro}. The task focuses on optimizing a Boiling Water Reactor (BWR) fuel lattice based on a GE-14–like assembly configuration. Design variables include uranium-235 enrichment levels for multiple fuel pins and gadolinium concentrations for burnable poison pins arranged within a symmetric lattice geometry. The optimization objective is multi-objective in nature, targeting a beginning-of-cycle infinite multiplication factor ($k_{\text{inf}}$) close to a prescribed value while maintaining the pin power peaking factor (PPF) below a safety threshold.

\begin{figure}[h!]
    \centering
    \includegraphics[width=0.4\linewidth]{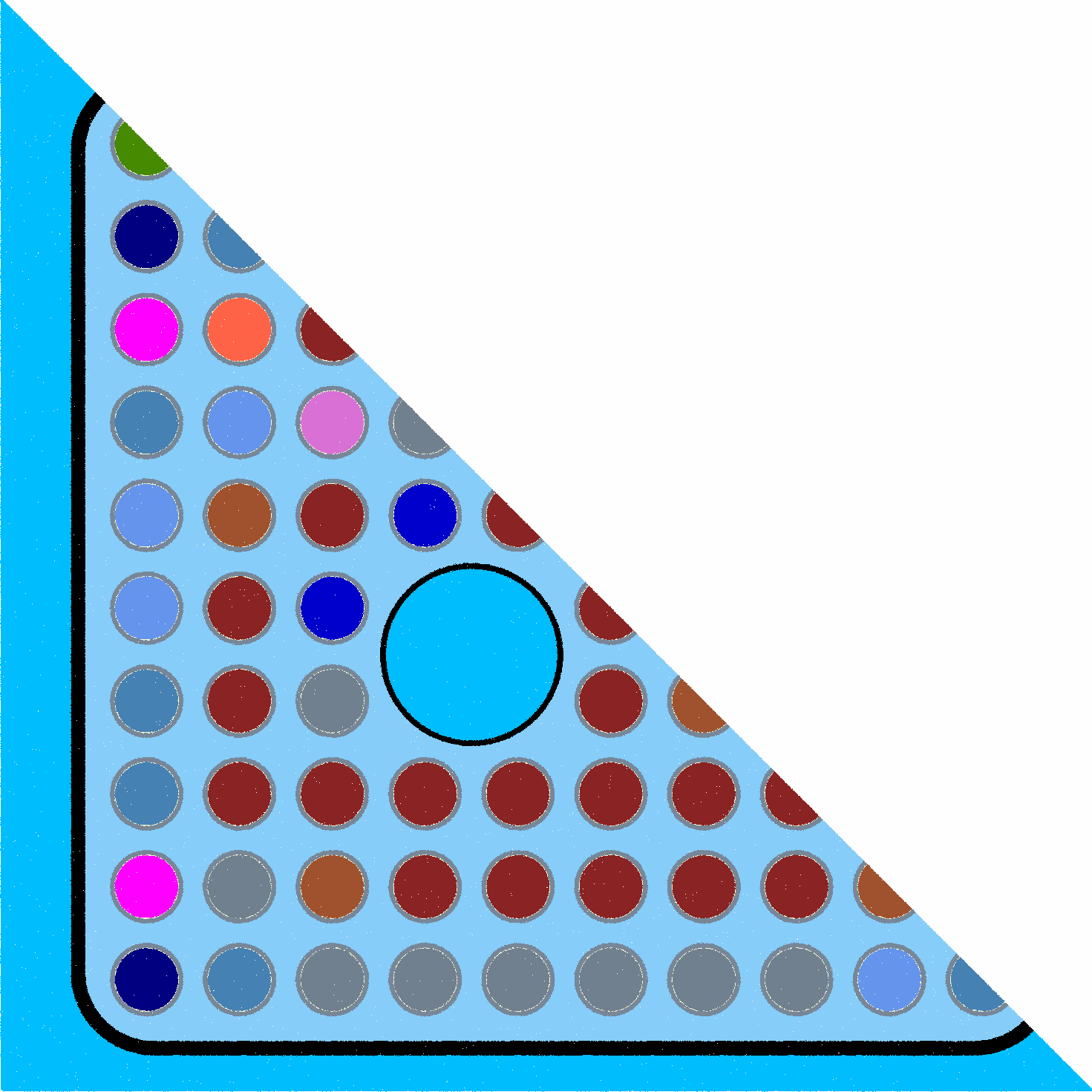}
    \caption{Half symmetry of GE-14 Dominant (DOM) fuel lattice. The image is generated by the authors using Casmo-5 software.}
    \label{fig:GE-14}
\end{figure}

The fuel lattice layout and pin-type arrangement are illustrated in Figure~\ref{fig:GE-14}. Candidate solutions are evaluated using high-fidelity neutronics simulations performed with CASMO-5, which computes lattice-level neutronic parameters for each proposed configuration. Constraint handling is embedded directly into the objective function through penalty terms that discourage deviations from the target $k_{\text{inf}}$ and violations of the PPF limit.

Within the LLMize framework, this problem is formulated as a black-box optimization task defined by a natural language problem description and an external evaluator. Optimization is performed using Optimization by Prompting (OPRO). At each iteration, the large language model proposes candidate fuel lattice configurations based on a meta-prompt containing the reactor physics objectives, operational constraints, and a history of previously evaluated solutions.

The optimization workflow combines LLM-based proposal generation with CASMO-5 evaluation and iterative feedback. This formulation allows reactor physics knowledge, safety considerations, and design heuristics to be injected directly in English. Through in-context learning, the language model infers relationships between enrichment distributions, gadolinium pin placement, reactivity control, and power peaking behavior from historical solution–score pairs.

\begin{figure}[h!]
    \centering
    \includegraphics[width=1\linewidth]{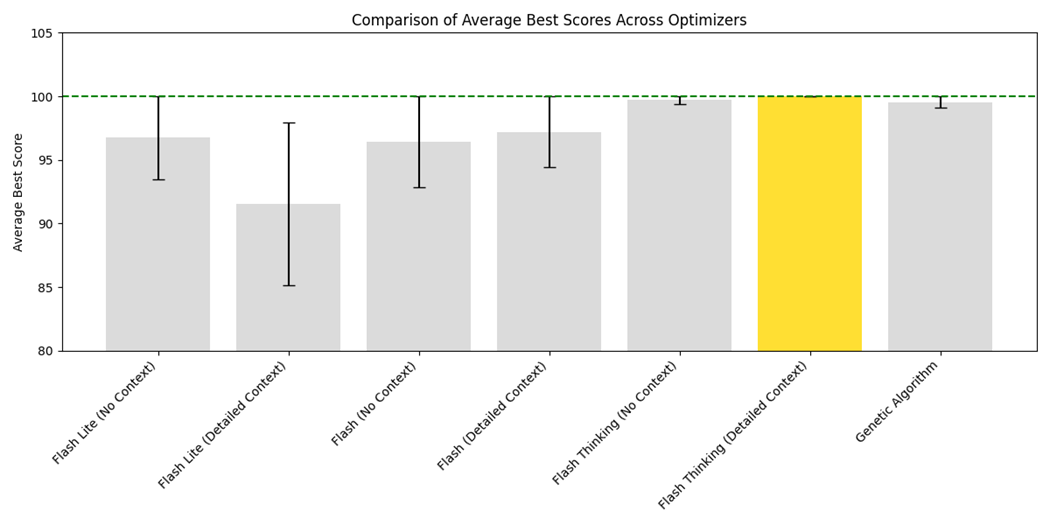}
    \caption{Comparison of Average Best Scores Across Optimizers for Different Prompting Strategies. The yellow bar highlights the optimal approach.}
    \label{fig:score_comparison}
\end{figure}

\textbf{Results.}  
As demonstrated in \citep{oktavian2024nuclearopro}, LLM-based optimization achieves performance that matches or exceeds a reference genetic algorithm across multiple optimization trials. Figure~\ref{fig:score_comparison} compares average best scores obtained using different LLM variants and prompting strategies, showing that reasoning-capable models with detailed context prompting consistently achieve near-optimal solutions. Convergence behavior indicates that LLM-driven optimization can effectively navigate the highly constrained and discrete design space of nuclear fuel lattices while maintaining physically acceptable solutions.

\section{Discussion and Future Work}

The results presented in this work illustrate both the strengths and limitations of LLM-based optimization within the LLMize framework. A central advantage of this approach is the ability to inject constraints, rules, and domain-specific knowledge directly through natural language. Rather than requiring practitioners to formalize objectives and constraints using specialized mathematical formulations, LLMize allows problem requirements, safety limits, and heuristic guidance to be expressed in plain English and enforced through an external evaluator. This design is particularly valuable in complex engineering and scientific domains, where expert knowledge is often qualitative and difficult to encode as explicit optimization operators.

The case studies also highlight an important tradeoff between optimization flexibility and computational efficiency. LLM-based optimization introduces nontrivial inference overhead compared to classical solvers, as each iteration requires one or more language model queries. For simple or well-structured problems such as low-dimensional convex optimization or linear programming, this overhead outweighs any practical benefit. In such settings, classical methods with access to gradients, closed-form solutions, or highly optimized solvers remain vastly more efficient and reliable. In these cases, LLMize primarily serves as a demonstration of capability rather than a competitive alternative.

In contrast, LLMize is most effective for complex, domain-specific optimization problems where constraints are highly structured, objectives are simulator-based, and problem formulations evolve over time. In these scenarios, the cost of language model inference is often small relative to the expense of objective evaluation, such as physics-based simulations or training machine learning models. Moreover, the ability to express domain knowledge, safety rules, and design heuristics in natural language can significantly reduce development time and lower the barrier to applying optimization techniques. The nuclear fuel lattice optimization case study illustrates this advantage clearly, as reactor physics constraints and operational considerations can be described naturally without requiring bespoke evolutionary operators or surrogate models.

Another strength of LLMize lies in its simplicity of setup. Defining a new optimization problem requires only a textual problem description, a black-box objective function, and minimal parsing logic to convert model outputs into candidate solutions. No algorithm-specific tuning or reformulation is necessary. This makes LLMize particularly attractive for practitioners who possess deep domain expertise but limited experience with advanced metaheuristic optimization methods. By shifting complexity from algorithm design to prompt design, LLMize enables rapid prototyping and experimentation across a wide range of problem domains.

Despite these advantages, several challenges remain. Scaling LLM-based optimization to high-dimensional problems introduces practical limitations related to prompt length, output formatting, and constraint enforcement. As the number of decision variables grows, representing solutions and historical evaluations in text form becomes increasingly cumbersome and may exceed the context limits of current language models. Complex prompts also increase the risk of malformed or infeasible outputs, necessitating more robust parsing, validation, and repair mechanisms.

Cost and efficiency considerations also become more pronounced at scale. While LLMize is well suited for low- to moderate-budget optimization tasks, extensive exploration in high-dimensional spaces may require a large number of language model queries, leading to increased runtime and cost. Careful batching, early stopping, and hybrid strategies are therefore essential to maintain practical efficiency.

Several promising directions for future work emerge from these observations. One direction is the development of hybrid optimization workflows that combine LLMize with classical methods, such as using LLMs to generate high-quality initial solutions or constraint-aware perturbations that are subsequently refined by local or gradient-based optimizers. Another direction involves fine-tuning or adapting language models to specific engineering domains, which may improve consistency, reduce hallucinations, and decrease reliance on extensive prompt engineering. Advances in structured output generation and constraint-aware decoding may further improve reliability, particularly for safety-critical applications.

Overall, LLMize demonstrates that large language models can function as flexible black-box optimizers when guided by in-context learning and external evaluation. While not a replacement for classical optimization methods in well-structured settings, LLMize provides a practical and accessible alternative for complex problems where constraints, heuristics, and expert intuition are most naturally expressed in language.

\section*{Acknowledgements}
This research made use of the resources of the High-Performance Computing Center at Idaho National Laboratory, which is supported by the Office of Nuclear Energy of the U.S. Department of Energy and the Nuclear Science User Facilities under Contract No. DE-AC07-05ID14517. This work also made use of Purdue University's Anvil computing cluster \citep{Song2022}.

\bibliographystyle{unsrtnat}
\bibliography{references}

\newpage

\end{document}